\documentclass[letterpaper, 10 pt, conference]{ieeeconf}  % Comment this line out if you need a4paper

\IEEEoverridecommandlockouts                              % This command is only needed if 
\usepackage{pifont}

\usepackage{xcolor}
\usepackage{adjustbox}
\usepackage{siunitx}
\usepackage{graphicx}
\usepackage{subcaption}
\usepackage{amsmath}
\usepackage{amssymb}
\usepackage{booktabs}
\usepackage{mathtools}

\usepackage{dsfont}

\usepackage{comment}
\usepackage{acronym}
\usepackage{multirow}
\usepackage{nicefrac}
\usepackage{times}
\usepackage{epsfig}
\usepackage{tabularx}
\usepackage{makecell}  % for \makecell
\usepackage[table]{xcolor}  % for \rowcolor with shades
\usepackage{float}
\makeatletter
\@ifundefined{labelindent}{}{%
  \let\labelindent\relax
}
\makeatother
\usepackage[inline,shortlabels]{enumitem}
\usepackage{afterpage}
\usepackage{diagbox}
\usepackage{colortbl}
\usepackage[font=small,labelfont=bf,compatibility=false]{caption}

\usepackage{threeparttable}
\usepackage{soul}
\usepackage[colorlinks=true, linkcolor=cyan, urlcolor=magenta]{hyperref}
\usepackage[flushleft]{threeparttablex}

\def\BibTeX{{\rm B\kern-.05em{\sc i\kern-.025em b}\kern-.08em
    T\kern-.1667em\lower.7ex\hbox{E}\kern-.125emX}}

\title{\LARGE \bf HetroD: A High-Fidelity Drone Dataset and Benchmark for Autonomous Driving in Heterogeneous Traffic}

\author{
\makebox[\linewidth][c]{%
Yu-Hsiang Chen$^{1,2}$\quad
Wei-Jer Chang$^{2}$\quad
Christian Kotulla$^{3}$\quad
Thomas Keutgens$^{3}$\quad
Steffen Runde$^{3}$
}
\and
\makebox[\linewidth][c]{%
Tobias Moers$^{3}$\quad
Christoph Klas$^{3}$\quad
Wei Zhan$^{2}$\quad
Masayoshi Tomizuka$^{2}$\quad
Yi-Ting Chen$^{1,\dagger}$
}\\[0.5em]
\makebox[\linewidth][c]{%
$^{1}$National Yang Ming Chiao Tung University \quad
$^{2}$UC Berkeley \quad
$^{3}$fka GmbH
}\\[0.3em]
\makebox[\linewidth][c]{%
† Corresponding Author
}
}

\begin{document}

\twocolumn[{%
\renewcommand\twocolumn[1][]{#1}  % 讓 \maketitle 不打斷兩欄
\maketitle
\thispagestyle{empty}
\pagestyle{empty}

\begin{center}
    \captionsetup{type=figure}
    \includegraphics[width=0.9\linewidth]{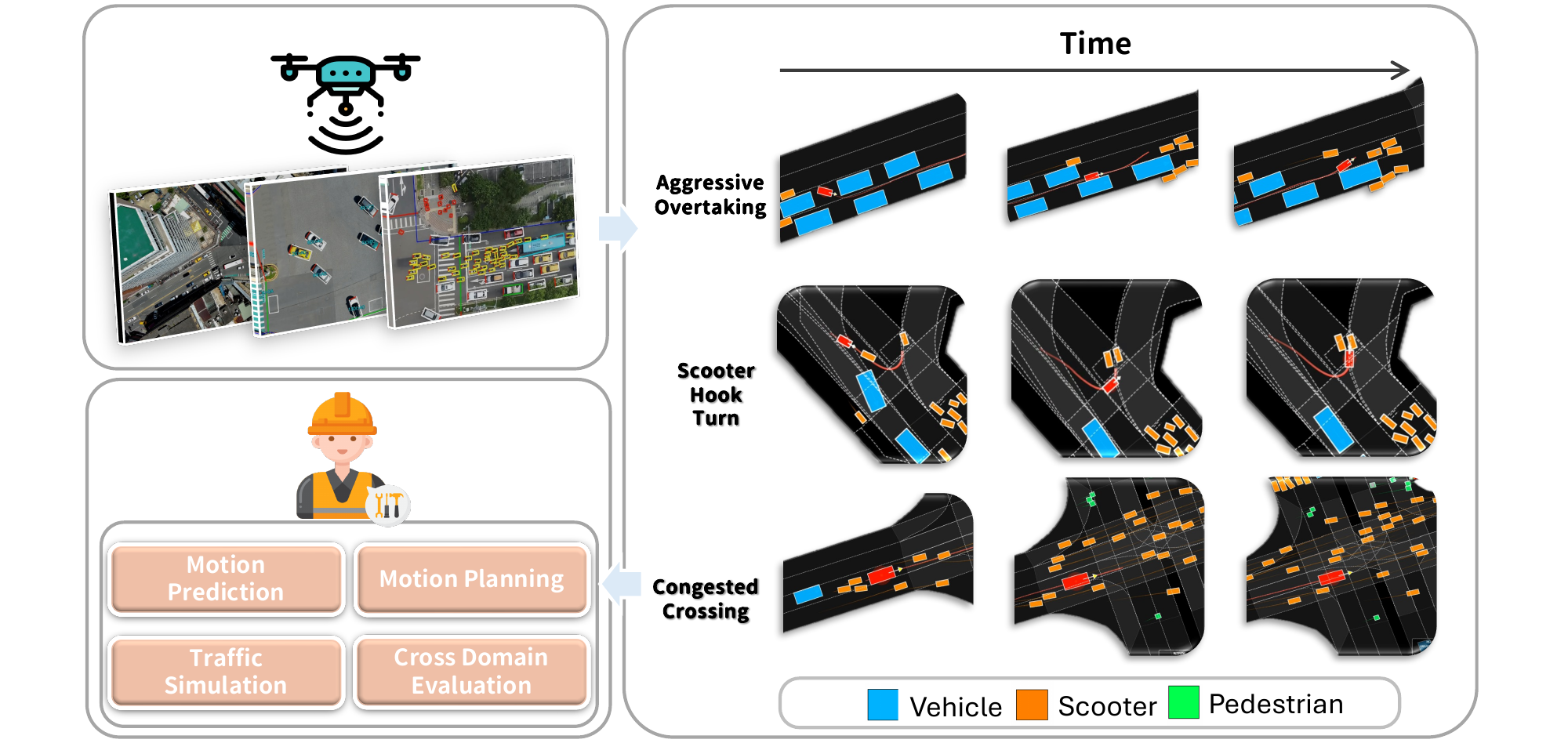}
    \captionof{figure}{\small
    \textit{HetroD} is a high-fidelity, drone dataset that captures unstructured maneuvers such as hook turns, aggressive overtakes, queue cutting, and congested crossings among vehicles, scooters, and pedestrians in heterogeneous traffic environments. 
    These maneuvers are critical for testing autonomous driving systems yet remain underexplored in the community.
    To address this, we construct a benchmark to evaluate existing methods in motion planning, motion prediction, traffic simulation, and conduct a thorough investigation of their generalization across datasets. 
    %\ychen{pls consider if you would like to keep the term "culturally grounded" in this article. if you could spell out its difference in your experiments, you can keep it.} 
    %
    % a cross-dataset evaluation of model performance
    %
    % The benchmark focuses on heterogeneous traffic environments 
    } 

    \label{fig:teaser}
\end{center}
}]

\begin{abstract}
We present HetroD, a dataset and benchmark for developing autonomous driving systems in heterogeneous environments. HetroD targets the critical challenge of navigating real-world heterogeneous traffic dominated by vulnerable road users (VRUs), including pedestrians, cyclists, and motorcyclists that interact with vehicles.
%
% \ychen{the first sentence is a good example. 1. it is too long. 2) "targeting..." i suggest to rewrite it as "HetroD targets at the critical challenge ..." } 
%
These mixed agent types exhibit complex behaviors such as hook turns, lane splitting, and informal right-of-way negotiation. Such behaviors pose significant challenges for autonomous vehicles but remain underrepresented in existing datasets focused on structured, lane-disciplined traffic. 
%
% \ychen{same as this sentence. you make it shorter and crisp. moreover, pls keep in mind if you would like to keep "culturally grounded" in this article.}
%
% \st{To collect data on these complex interactions among different agent types, we conduct drone-based data collection in urban environments.}
%
% \ychen{capture is a strange term here. do you mean "collecting data" or "modeling interaction"} 
%
%
%\ychen{tracking, what does it mean?} 
%
To bridge the gap, we collect a large-scale drone-based dataset to provide a holistic observation of traffic scenes with centimeter-accurate annotations, HD maps, and traffic signal states.
%
% \st{Through intensive manual annotation and rigorous quality control,} \ychen{this sentence is basically the same the next sentence. you did not provide much new information.} 
%
% \ychen{this sentence is strange. who is the subject? it is we employ tracking..., we provide..., }
%
% \st{We provide centimeter-accurate annotations, HD maps, and traffic signal states.} 
%
% \st{Additionally, we develop a modular toolchain for extracting per-agent scenarios.}
We further develop a modular toolkit for extracting per-agent scenarios to support downstream task development.
% \ychen{i suggest to use another sentence to describe you provide a toolchain.}
%
In total, the dataset comprises over 65.4k high-fidelity agent trajectories, ~70\% of which are from VRUs.
%\ychen{maybe highlight trajectories of different agent types?}
%
HetroD supports modeling of VRU behaviors in dense, heterogeneous traffic and provides standardized benchmarks for forecasting, planning, and simulation tasks. 
Evaluation results reveal that state-of-the-art prediction and planning models
%
% \ychen{only AD models?} yh: changed to be more specific
%
struggle with the challenges presented by our dataset: they fail to predict lateral VRU movements, cannot handle unstructured maneuvers, and exhibit limited performance in dense and multi-agent scenarios, highlighting the need for more robust approaches to heterogeneous traffic.
See our project page for more examples: \href{https://hetroddata.github.io/HetroD/}{https://hetroddata.github.io/HetroD/}
\end{abstract}    
\section{Introduction}

\begin{table*}[t!]
  \centering
  \setlength{\tabcolsep}{4pt}
  \begin{threeparttable}
    \caption{
      \textbf{Comparison of Datasets on Interaction, Density \& Diversity Metrics.}
      We report key statistics across on-board and drone-view datasets. \textit{Interaction Scale}\tnote{1} is the total number of interactions, computed per dataset by aggregating over all its scenarios and then normalized across datasets. \textit{Heterogeneous Interaction Scale}\tnote{2} counts cross-type interactions with the same per-dataset aggregation and normalization. \textit{Geographical Density}\tnote{3} represents the average number of agents per unit area $A$ within an 8-second window. \textit{VRUs}\tnote{4} denotes the proportion of VRUs among all traffic agents. All metrics except VRUs are min–max normalized to [0,1] across datasets where the metric is available; VRUs is reported in percent. Normalization is used for cross-dataset comparability rather than absolute interaction frequency. Boldface indicates the highest value among all datasets, while underlined values denote the highest among drone-view datasets.
      % \ychen{the notions of metrics listed below the table should be linked to the texts in the table and caption.}
      % \textcolor{blue}{add number link to the table and caption}
      }
    \label{tab:datasets_comparison}
    \begin{tabular*}{\textwidth}{@{\extracolsep{\fill}} l l c c c c c c @{}}
      \toprule
      \textbf{Dataset} &
      \makecell{\textbf{Platform}} &
      \makecell{\textbf{Tracks}} &
      \makecell{\textbf{Duration}} &
      \makecell{\textbf{Interaction}\\\textbf{Scale}\tnote{1}} &
      \makecell{\textbf{Heterogeneous}\\\textbf{Interaction Scale}\tnote{2}} &
      \makecell{\textbf{Geographical}\\\textbf{Density}\tnote{3}} &
      \makecell{\textbf{VRUs}\\\textbf{(\%)}\tnote{4}} \\
      \midrule
      NuScenes~\cite{nuScenes}      & On-board & $\sim$90k$^\dagger$ & 320h   & 0.675 & 0.549 & —     & 20.1\%      \\
      Waymo~\cite{Waymo}            & On-board & 7.6M               & 574h   & 1.000 & 1.000 & —     & 11.5\%      \\
      Argoverse2~\cite{Argoverse2}  & On-board & 13.9M              & 763h   & 0.632 & 0.318 & —     & 10.0\%      \\
      NuPlan~\cite{nuplan}          & On-board & $\sim$5M$^\dagger$ & 1282h  & 0.274 & 0.213 & —     & 46.3\%      \\
      \addlinespace
      INTERACTION~\cite{INTERACTION}& Drone    & 40k                & 16.5h  & 0.132 & —     & 0.011 & —      \\
      inD~\cite{inD}                & Drone    & 13.5k              & 10h    & 0.071 & 0.185 & 0.023 & 39.4\% \\
      SinD~\cite{SIND}              & Drone    & 13.2k              & 7.02h  & 0.099 & 0.324 & 0.016 & 62.1\% \\
      \rowcolor{white}
      \textbf{HetroD}               & Drone    & \underline{65.4k}  & \underline{17.5h}
                                    & \underline{0.223} & \underline{0.889}
                                    & \textbf{0.026} & \textbf{69.9\%} \\
      \bottomrule
    \end{tabular*}
    \begin{tablenotes}[flushleft]\footnotesize
  % Two-column layout for footnotes
  \begin{minipage}[t]{0.48\textwidth}
    \item[$\dagger$] Estimated values based on official statistics.
    \item[—] Metric not available.
    \item[1] $\mathcal{S}_{\text{inter}} = \sum_{\mathrm{scenarios}} \mathcal{D}_{\text{inter}}$.
    \item[2] $\mathcal{S}_{\text{het}} = \sum_{\mathrm{scenarios}} \sum_{i,j} 1_{(\mathrm{TTC}_{i,j}<2\,\mathrm{s} \land \mathrm{type}_i\neq\mathrm{type}_j)}$.
  \end{minipage}%
  \hfill
  \begin{minipage}[t]{0.48\textwidth}
    \item[3] $\mathcal{D}_{\text{geo}} = N / A$, where $N$ is the number of agents within an 8\,s window and $A$ is the corresponding area.
    \item[4] $\mathrm{VRUs} = 100 \times \frac{N_{\text{VRU}}}{N_{\text{VRU}}+N_{\text{Veh}}}$ 
    (VRU: pedestrians, bicycles/cyclists, motorcycles, tricycles; Vehicles: cars, trucks, buses, vans).

  \end{minipage}
\end{tablenotes}
  \end{threeparttable}
\end{table*}

%Suggestion:introduction can start in 1) heterogeious traffic are important problem to solve and is high priority (cyclusts, pedestrians). 2)In Recent years, data-driven modeling has become more popular,...., however, the datasets mainly consists of veh2veh interactions(refer to Table2) 3)This result in most simulators model as log-replay,...

%One of the most challenging mixed-agent traffic env. Hihlgiht 1)Cutural difference 2)high hetorgenious interaction scale 3) and why we use drone , to create high-quality,,--> this provide a new

%Describe what we did 1) How we collect data 2)unified tooling 3)-->which enabled Detail analysis prediction/planning performance ,briefly mention the takeway of the results

%conrbitution

% yh: im wondering if the third part is overlapping with the contribution part? 

% \ychen{what is the message you would like to convey for this part "increasingly share road space with vehicles in dense urban centers worldwide"? Is it necessary to keep these words?}
% yh: the message had been shortend
Navigating heterogeneous traffic remains one of the core challenges in the development of autonomous driving systems, particularly due to vulnerable road users (VRUs), including cyclists, pedestrians, and motorcyclists, who interact with vehicles in complex ways. 
In recent years, data-driven modeling has become the dominant approach for autonomous driving development, as it provides a scalable way to capture complex traffic interactions with less human effort than rule-based modeling. 
However, most publicly available datasets primarily capture lane-disciplined traffic and vehicle-to-vehicle interactions~\cite{surveyondata}. 
%
% Because most datasets focus on vehicles, 
%
They include little data on VRUs (Table~\ref{tab:datasets_comparison}) and heterogeneous interactions. 
As a result, they miss many real-world situations where different road users compete for space and negotiate right-of-way through subtle, culture-specific cues~\cite{TraPHic}. 
Therefore, downstream models and widely used simulators inherit these biases: they either hard-code simplified VRU templates~\cite{carla+} or merely replay recorded trajectories, such as Waymax~\cite{waymax} and NuPlan~\cite{nuplan}, which limits their ability to capture heterogeneous reactive dynamics. These limitations are further discussed in Section~\ref{sec:2_related_work}.

%this part probably should refer to some of the table 1 statistics, and directly mention how HetroD is different than SinD

%I think we should also highlight why using drone view instead of on-board modules, drone view provide more accurate, longer observations?

This gap between current datasets and real-world scenes calls for data that captures the intricate interactions among mixed agents. 
Achieving such comprehensive data collection requires observation methods that overcome the occlusions and limited field-of-view inherent in on-board sensors. 
Drone-based observation provides a holistic scene coverage and temporal evolution of traffic participants, essential attributes for VRU interaction modeling. 

We introduce \textit{HetroD}, a drone-captured dataset collected across six topologically diverse, high-traffic urban locations in Taiwan.
While the existing drone-based dataset SinD~\cite{SIND} marks an important first step in capturing heterogeneous traffic interactions, HetroD offers a substantially larger interaction scale, with up to twice the number of cross-agent interactions (Table~\ref{tab:datasets_comparison}).
%
% \st{Existing drone-based dataset SIND~\cite{SIND} initiate the first attempt providing a good dataset for exploring the interaction modeling for heterogenous traffic. 
% However, HetroD offers significantly higher interaction scale with up to two-fold higher cross-agent interaction counts (Table~\ref{tab:datasets_comparison}).}
%
In addition, HetroD involves a wide range of intricate maneuvers such as hook turns, lane splitting, and aggressive overtakes and offers topological diversity across six intersection archetypes.
The dataset further includes centimeter-accurate HD maps, bounding boxes, and traffic signal states.
%
%\ychen{they do not have them?}
% yh: i think their motion had high randomness, not like hetrod
%
% Moreover, the dataset contains topological breadth spanning six intersection archetypes
%
% with centimeter-accurate HD maps, bounding boxes, and traffic signal states. 
%
Together, these traits position HetroD as a new testbed for developing autonomous driving systems in dense heterogeneous traffic.

\noindent Our contributions are summarized as follows:
\begin{itemize}
    \item We construct a drone dataset with centimeter-level annotations of heterogeneous traffic. The dataset spans 17.5 hours and contains over 65.4k agent tracks, 70\% from VRUs.  
    \item We establish benchmarks with standardized evaluation protocols for motion prediction, planning, and cross-dataset evaluation.
    %We build a benchmark suite comprising heterogeneous scenarios, baseline tasks, and a plug-and-play conversion toolchain. 
    %
    % \ychen{1. you do not need to specify again it contains heterogeneous scenarios, as the previous sentence has mentioned it. 2. what does it mean baseline tasks? 3. it is hard to understand what does it mean "plug-and-play conversion toolchain," without providing any context in the introduction.}
    \item We show that state-of-the-art prediction and planning methods face clear limitations on HetroD: they struggle to predict lateral VRU movements, handle specific maneuvers, and sustain performance in dense multi-agent scenarios.
    % \item We report evidence that HetroD reveals common failure modes of current state-of-the-art prediction and planning methods in heterogeneous traffic environments.
    % \ychen{Can you be more specific -- what are the potential causes of these failure modes?}
    
\end{itemize}

\section{Related Work} \label{sec:2_related_work}

Autonomous driving datasets vary by their sensing modalities and deployment context. We group related work into four categories: on-board, infrastructure-view, drone-view datasets, and unified development frameworks. Table~\ref{tab:datasets_comparison} summarizes key characteristics of relevant datasets.
%
% \ychen{We provide a summary of relevant datasets in Table~\ref{tab:datasets_comparison.}}

\noindent\textbf{On-Board Sensor Datasets.} Prior works~\cite{KITTI,h3d,Waymo,nuScenes,lyft,bdd100k,pandaset,Argoverse2,MONA,AD4CHE,WOMD} offer rich multimodal data but suffer from occlusions and limited VRU coverage in dense traffic~\cite{KITTI,Waymo,nuScenes}. While METEOR~\cite{chandra2022meteoradenseheterogeneousunstructured} pioneered heterogeneous traffic capture, its vehicle-centric data collection approach underrepresents VRU interactions. Furthermore, it lacks HD maps and comprehensive annotations needed for detailed traffic analysis.
% \ychen{this sentence sounds strange. why lacking hd maps and comprehensive annotations lead to vehicle-centric data?}

\noindent\textbf{Infrastructure-View Datasets.} These datasets~\cite{NGSIM,CITR_DUT,dairv2x,opv2v,V2X-Seq,V2AIX,v2xreal,tumtraf,Accid3nD} use fixed cameras or V2X sensors to reduce occlusion, but often suffer from low spatial resolution that hampers small object detection~\cite{NGSIM,V2X-Seq}, inaccurate camera calibration affecting accurate localization~\cite{v2xreal,tumtraf}, or limited cross-agent type diversity~\cite{V2AIX,dairv2x}, limiting their utility for modeling heterogeneous agent behaviors.
% \ychen{what does it mean lack resolution? lack calibration, what does it mean? class diversity ==> cross-agent type diversity?}

\noindent\textbf{Drone-View Datasets.} Existing drone datasets~\cite{Stanford_Drone,highD,INTERACTION,CITR_DUT,rounD,openDD,inD,Automatum,exID,SIND,JKU_DORA,CTV,CitySim,GroundMix,dsc3d} provide occlusion-free, global views and are ideal for interaction modeling and analysis. However, many are collected in lane-disciplined settings~\cite{highD,Automatum} and exhibit fragmented VRU tracks due to the difficulty of annotating small objects~\cite{SDDComplex,dsc3d}.
%
% tracking limitations. 
%
In addition, they also underrepresent unstructured maneuvers, such as informal yielding, weaving, or reverse flows. 
To the best of our knowledge, we are the first dataset 
%
% no existing public drone dataset 
%
that provides both per agent, centimeter accuracy in ground truth, and wide area coverage across diverse, heterogeneous urban environments for benchmarking autonomous vehicles in VRU-rich contexts.
% limiting their applicability to safety critical or VRUs-aware tasks.
% \ychen{please check my modifications.}
% \textcolor{blue}{checked! i learned a  lot }

\noindent\textbf{Unified Development Toolkit.} Recent development toolkits for tasks such as trajectory forecasting, scenario generation, and reinforcement learning~\cite{metadrive,scenarionet,trajdata,unitraj,charraut2025vmaxreinforcementlearningframework,gpudrive,dronalize} offer standardized interfaces to facilitate rapid development.
%
% rely on high-quality upstream data (especially fine-grained, interaction-aware trajectories and agent-centric scenarios).
%
%
However, existing drone datasets lack a suitable toolkit to enable collective development within the community.
 To address this, we make the first attempt to streamline existing toolkits such as ScenarioNet~\cite{scenarionet} and GPUDrive~\cite{gpudrive} (see Fig.~\ref{fig:tool-framework}) to be compatible with HetroD.
We invite the community to tackle these critical challenges collectively.
%
% \ychen{please check.}
% \textcolor{blue}{checked! i learned a  lot }
%
%
% the toolchains required to produce such structured annotations remain underdeveloped in existing drone datasets, limiting their utility for downstream tasks.

\begin{figure}[t!]
    \centering
    \includegraphics[width=0.48\textwidth]{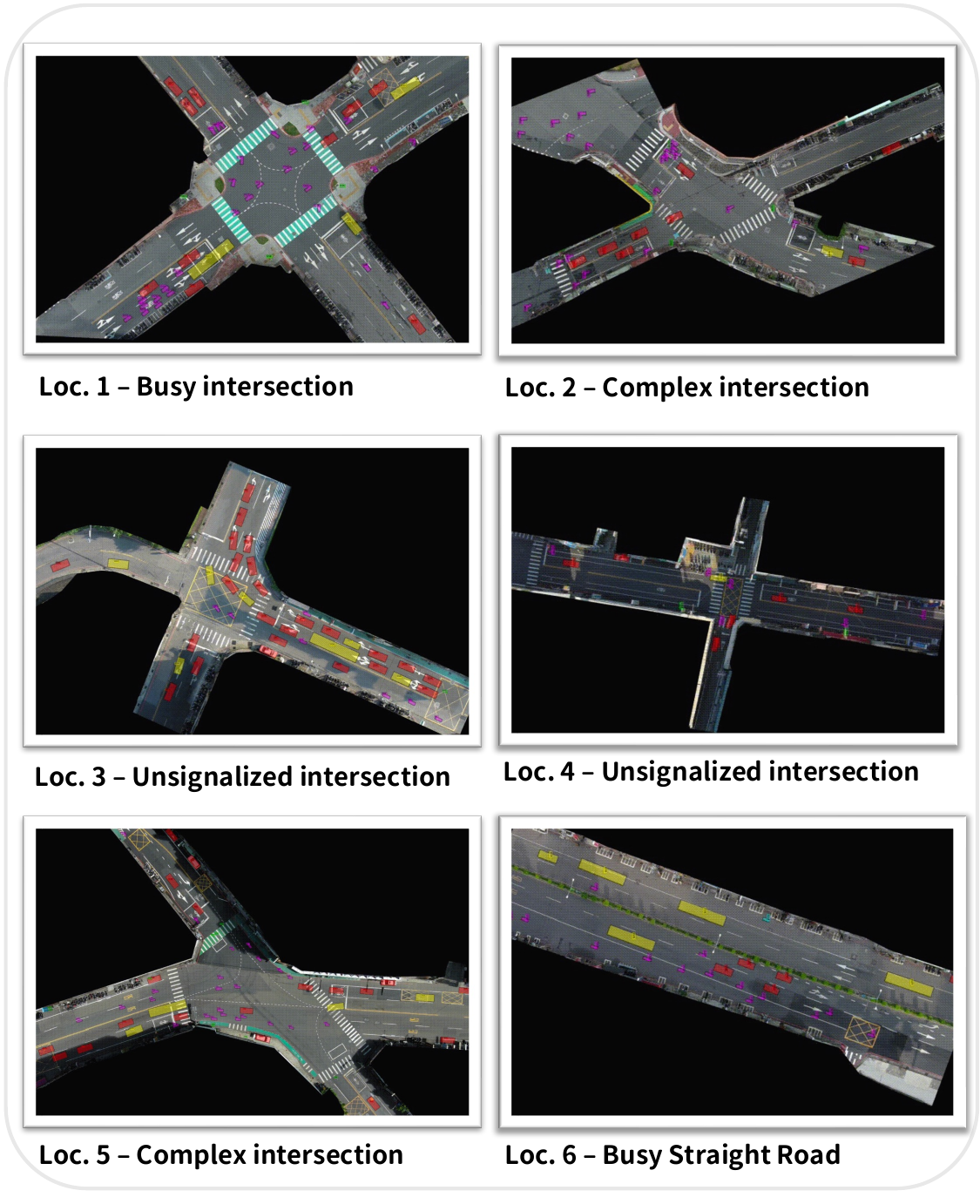}
    \caption{
    Aerial views of the six recording locations in HetroD, capturing diverse urban traffic scenarios.
    }
    \vspace{-10pt}
    \label{fig:location_image}

\end{figure}

% \noindent HetroD is designed to close these gaps by offering dense, heterogeneous scenarios with high-fidelity annotations, cultural motion diversity, and seamless framework compatibility (see Fig.~\ref{fig:tool-framework}).

\section{The HetroD Dataset}

\textit{HetroD} is a large-scale drone-view dataset comprising 17.5 hours of ultra-high-resolution (5.4K) video, collected across six topologically and behaviorally distinct urban sites in Taiwan. As shown in Fig.~\ref{fig:location_image}, the dataset captures diverse traffic scenarios including busy signalized intersections (Locations 1, 2, and 5), unsignalized intersections (Locations 3 and 4), and a busy straight road segment (Location 6). It encompasses over 65.4k unique trajectories across these varied traffic environments—archetypes rarely represented together in existing datasets.

\begin{figure}[t!]
    \centering
    \includegraphics[width=0.48\textwidth]{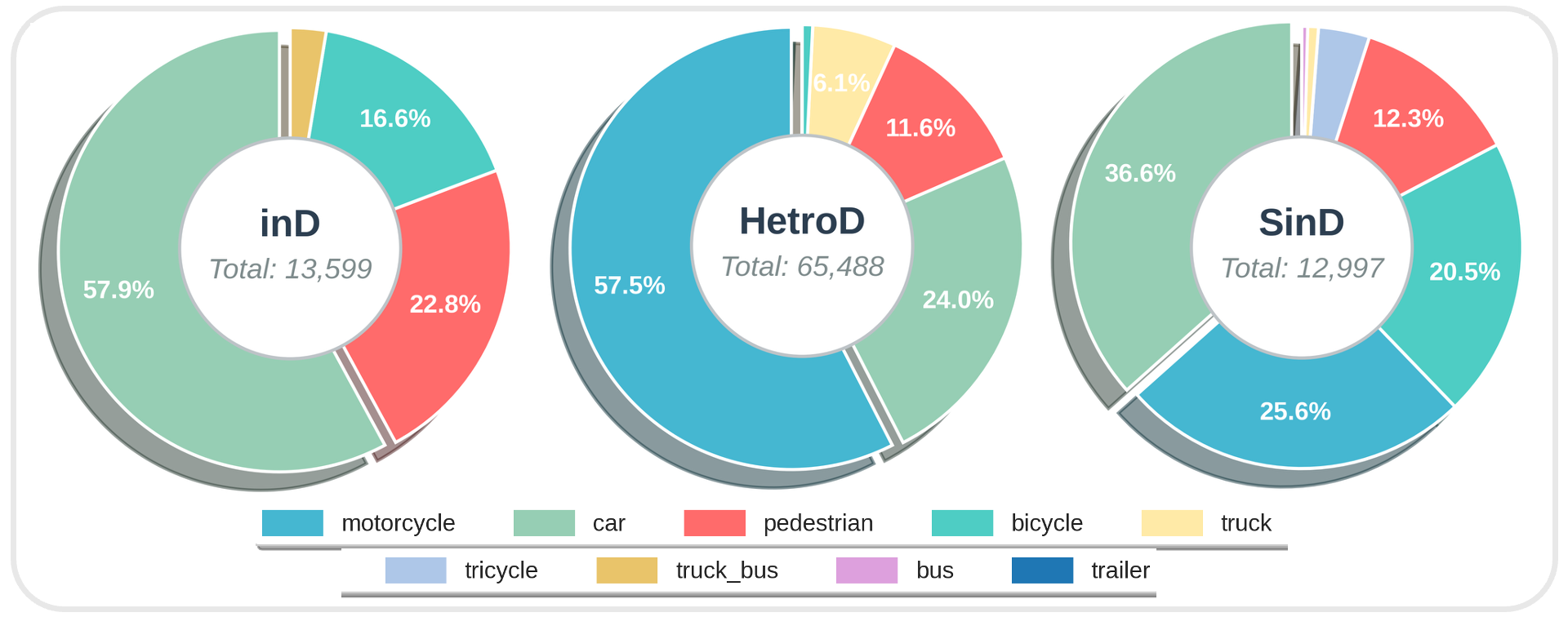}
    \caption{Agent-type distribution of HetroD (center) compared against two prior datasets, inD and SinD.}
    % \vspace{-10pt}
    \label{fig:agent_type}
\end{figure}

\begin{figure}[!t]
    \centering
    \includegraphics[width=0.48\textwidth]{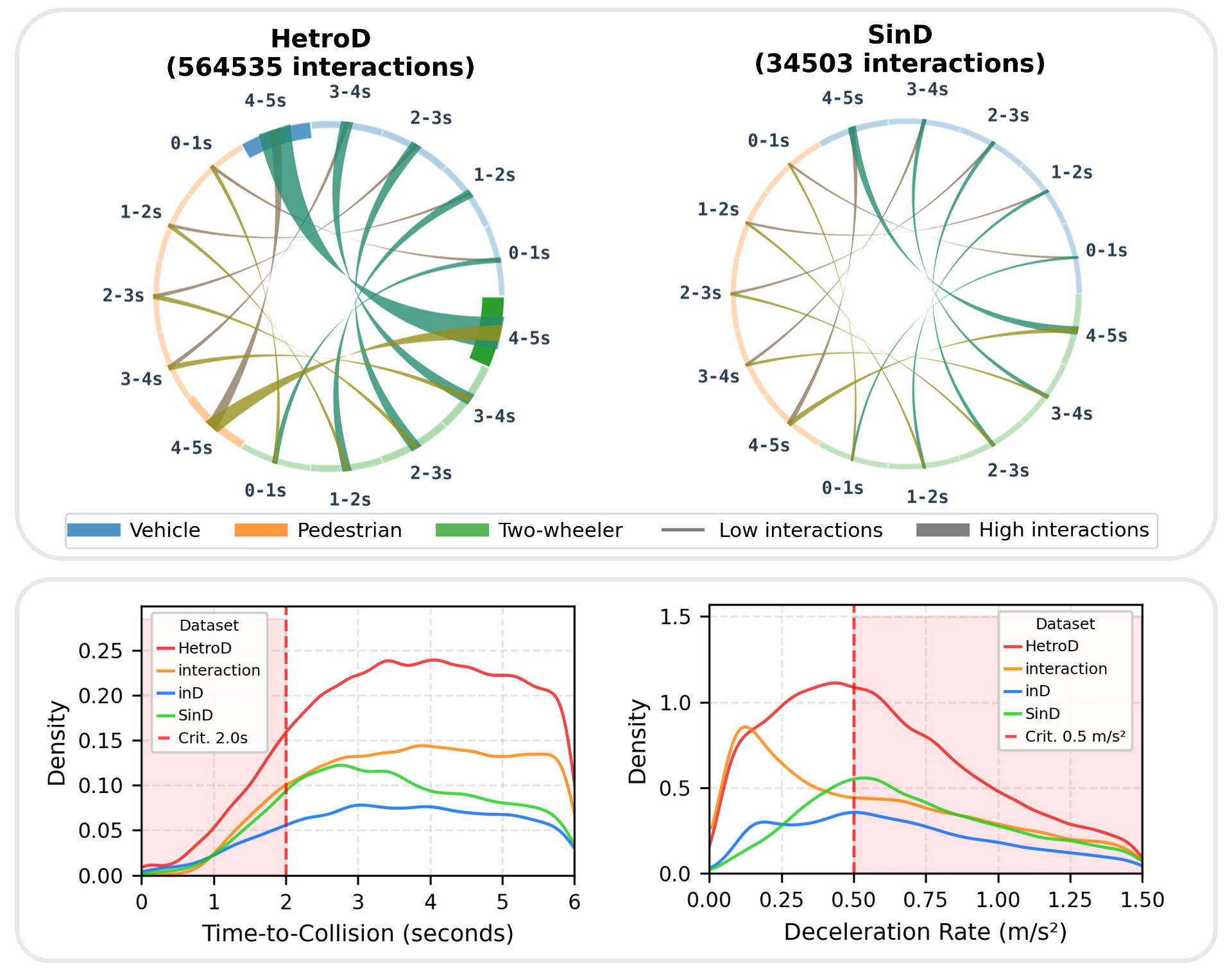}
    \caption{\textbf{Cross-type interaction patterns and TTC/DRAC distributions.}
    In the chord diagrams (top), link thickness indicates the number of cross-type interactions between agent categories, with color denoting time-to-collision (TTC)~\cite{ttc} bands (0–1, 1–2, 2–3, 3–4, 4–5\,s; lower indicates higher risk). The bottom panels show marginal distributions of TTC and deceleration rate to avoid crash (DRAC)~\cite{drac}. HetroD exhibits denser and riskier cross-type interactions, particularly among vehicles, two-wheelers, and pedestrians with a clear shift toward shorter TTC and higher DRAC, highlighting the prevalence of complex VRU interactions in HetroD compared to other datasets.}
    % \vspace{-10pt}
    \label{fig:chord}
\end{figure}

\subsection{Dataset Construction} \label{sec:dataset_construction}
HetroD is created from drone videos recorded continuously for approximately 20 minutes per flight at 25 or 30 FPS. Location 5 was recorded from 120 m above ground level (AGL), while all other locations were recorded from 100 m AGL. The videos are processed by an automated pipeline that removes distortion using the drone’s intrinsic camera parameters and stabilizes the videos. Afterwards, visible road users are detected and classified using deep neural networks. The subsequent tracking step creates trajectories from the detections, while Kalman filter~\cite{kalmen_filter} predictions are used to fill in small gaps. The trajectories are refined afterwards using a series of post-processing steps to create smooth and precise trajectories and to remove false positives. The refined trajectories then undergo two quality assurance cycles. In each cycle, the results of automatic checks and manual inspection are reviewed and fixed. This yields highly accurate trajectory data without tracking errors. For each recording location, HD maps are created from an orthophoto of the respective scene. We provide maps in Lanelet2 \cite{lanelet2} and OpenDRIVE~\cite{opendrive} formats. The data, orthophoto and HD map are referenced against a common local metric frame $\mathcal{F}_\text{location}$ for each location. Traffic light states are derived from ground-based recordings using deep neural networks or from traffic light state diagrams for the location. In case of ground-based recordings, a mask is applied to ensure privacy of road users. Trajectory and traffic light data are synchronized in time to enable matching of traffic light states to the trajectory data.

\begin{figure}[t!]
    \centering
    \includegraphics[width=0.48\textwidth]{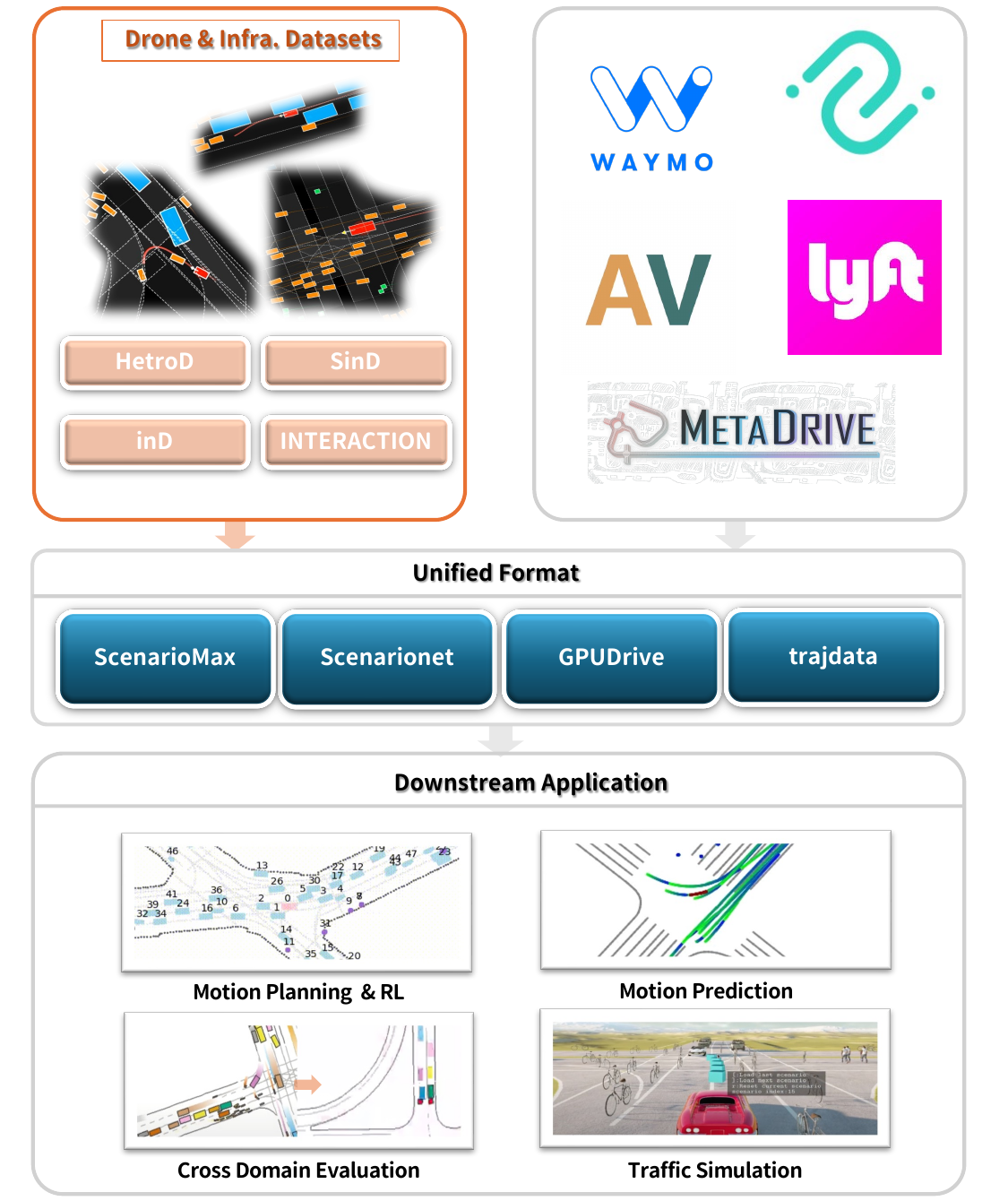}
    \caption{
    % HetroD features heterogeneous agents, enabling richer interactions and complex multi-agent and VRUs behavior modeling. 
    %
    We develop a unified development toolkit that converts a wide range of traffic scene datasets into standardized, agent-centric data formats~\cite{scenarionet,trajdata, charraut2025vmaxreinforcementlearningframework,gpudrive}, enabling seamless comparisons across datasets for forecasting, planning, and simulation. 
    % yh: Im a little confusd on how to arrange 3,4 and 5 cuz they are aligned with the order of the article now.
    % \ychen{I suggest moving Fig. 2 after Fig. 3, 4, 5. These figures are about the dataset. Fig 2 is a layer on top of the dataset. Thus, it should be discussed later. note: i felt the order of 3, 4, and 5 should be rearranged.}
    %
    }
    % \vspace{-10pt}
    \label{fig:tool-framework}

\end{figure}

\subsection{Dataset Properties}
\label{sec:dataset_Property}

To meaningfully quantify traffic complexity in dense, heterogeneous environments, we introduce four normalized metrics that capture spatial, behavioral, and interaction-level diversity (see Table~\ref{tab:datasets_comparison}). These metrics (ranging from \textit{interaction scale} to the presence of \textit{VRUs}) are computed based on full drone-captured datasets. To ensure fair comparison across datasets of varying durations, all metrics are upsampled or downsampled to match the duration of HetroD (17.5 hours), and normalized accordingly. Scale-related metrics are computed using full-dataset coverage. Together, these metrics enable robust cross-platform and cross-dataset comparisons of traffic complexity. Our dataset analysis reveals two fundamental findings:
\begin{itemize}
    \item \textbf{Dataset scale and interaction complexity:} HetroD contains the largest number of unique agent tracks and exhibits the highest levels of \textit{interaction scale} and \textit{heterogeneous interaction scale} among existing drone-view datasets.
    
    \item \textbf{Cultural and behavioral richness:} While SinD~\cite{SIND} offers a balanced distribution of agent types, HetroD presents a unique setting where \textit{scooters outnumber vehicles}, reflecting traffic patterns not captured in prior datasets (Fig.~\ref{fig:agent_type}). These agents demonstrate complex behaviors such as weaving, filtering, and informal negotiation, rarely modeled at scale. Risk indicators like TTC~\cite{ttc} and DRAC~\cite{drac} reveal significantly higher latent conflict rates (Fig.~\ref{fig:chord}).
    % \item \textbf{Modeling utility:} Experiments show substantial distributional shifts induced by rare maneuvers (e.g., informal U-turns, reverse flows), highlighting the need for datasets that capture such diversity.
\end{itemize}

\subsection{Unified Development Toolkit}
\label{sec:toolkit}
To facilitate community adoption, we develop a toolkit that converts HetroD data into formats compatible with popular autonomous driving frameworks (Fig.~\ref{fig:tool-framework}) including ScenarioNet~\cite{scenarionet}, ScenarioMax~\cite{charraut2025vmaxreinforcementlearningframework}, GPUDrive~\cite{gpudrive}, and trajdata~\cite{trajdata}. This enables researchers to leverage HetroD for motion prediction, planning, simulation, and cross-dataset evaluation without extensive preprocessing, making the dataset immediately usable within existing development workflows.

Leveraging this diversity, HetroD fills a long-standing gap in heterogeneous traffic modeling and unlocks two pivotal research axes:
\begin{itemize}
    \item \textbf{High-fidelity heterogeneous traffic simulation:} From full-scene replay to reactive VRUs modeling.
    
    \item \textbf{VRUs motion prediction and cross-domain generalization:} Enabling out-of-distribution testing on rare, unstructured maneuvers.
\end{itemize}

% % Full-width comparison table with explanations in caption and formulas in footnotes

\section{Evaluation}

We construct a set of challenging per-agent scenarios from HetroD. Specifically, we sample agents exhibiting non-trivial behavior such as long traversals, abrupt heading changes, and dense interactions within multi-agent contexts. These selected agents are used to instantiate per-agent scenarios for evaluation.

\subsection{Motion Forecasting} 

We evaluate the cross-dataset generalization of two state-of-the-art predictors, \textit{MTR}~\cite{mtr} and \textit{Wayformer}~\cite{wayformer}, on HetroD under the \textit{UniTraj}~\cite{unitraj} protocol. Models are trained on NuScenes, Waymo, SinD, and HetroD, and evaluated on each test set using the Brier-FDE metric~\cite{Argoverse2}, which measures probabilistic endpoint accuracy (lower is better). As shown in Table~\ref{tab:cross_dataset_main}, performance drops significantly, particularly when transferring from drone-view training data to on-board test sets. To further analyze the factors contributing to model performance, we follow the original MTR implementation, training with anchor files (trajectory endpoint cluster centers that guide the model's search space) generated from each training dataset and evaluate three distinct data-splitting regimes on HetroD, as summarized in Table~\ref{tab:hetrod_ablation}:
(1) a random split, where training and testing samples are randomly partitioned;
(2) a map-based split, with training and testing conducted on geographically disjoint locations; and
(3) a time-based split, where training uses data from busy hours (7-9 AM) and testing uses data from non-busy hours (11 AM-12 PM). Our findings are:
\begin{itemize}
    \item \textbf{In-domain superiority:} Training and testing on the same dataset yields strongest performance but exhibits large cross-domain gaps.

    \item \textbf{Drone-view generalization advantage:} Models trained on drone-view data generalize better to other drone datasets than on-board models generalize to drone-view tests, owing to the near-complete, occlusion-free coverage of drone videos, whereas on-board datasets often yield fragmented or discontinuous tracks for non-ego agents due to limited field of view (FOV) and occlusions. Notably, even when MTR is provided with test-domain anchors, the performance gap remains substantial, suggesting that viewpoint geometry rather than anchor mismatch drives this distribution shift.

    \item \textbf{Map sensitivity:} Map shifts substantially harm in-domain accuracy (166\% increase in error for MTR).
    
    \item \textbf{Temporal robustness:} Time shifts have minimal in-domain impact (5\% decrease in error for MTR).
    
    \item \textbf{Sensitivity to anchor priors:} In MTR, the low in-domain Brier-FDE rises sharply with map changes, indicating strong dependence on the anchor prior.
    
    \item \textbf{Model trade-offs:} MTR outperforms Wayformer when trained on drone data but is more sensitive to map shifts, while Wayformer remains more consistent across on-board datasets.
\end{itemize}

These results reveal fundamental limits of current forecasting models in heterogeneous traffic and highlight the value of occlusion-free drone data, which provides complete scene coverage and enables new research opportunities on dense multi-agent interactions.

\subsubsection{Scenario-Conditioned Evaluation}

To analyze failure modes, we stratify evaluation by agent type and cross-type interaction risk. Agent types are \textit{vehicles}, \textit{two-wheelers (scooters \& motorcycles)}, and \textit{pedestrians}. Cross-type interaction risk is measured by the minimum time-to-collision (TTC) between the focal agent and agents of different types: \textit{High} ($\text{TTC}<2$\,s), \textit{Moderate} ($2 \le \text{TTC}<4$\,s), and \textit{Low} ($\text{TTC}>4$\,s). We also examine \textit{scene-level heterogeneity} (low vs.\ high) using a combined score of local density and Shannon diversity~\cite{shannon} within a 10\,m radius of the ego agent, and bin scenes accordingly.   

Table~\ref{tab:scenario_conditioned} shows that denser, more heterogeneous scenes and higher cross-type interaction risk yield larger errors (High $>$ Moderate $>$ Low across training sets). Across agent types, two-wheelers are generally the hardest to forecast, while pedestrians exhibit the smallest \emph{absolute} Brier-FDE largely because they travel shorter distances. In terms of \emph{predictability of behavior}, vehicles are typically easier to model due to more regular lane-following dynamics, whereas pedestrians’ intentions can still be harder to anticipate despite their short displacements. Irregular two-wheeler maneuvers (e.g., weaving, filtering, abrupt lane changes) further increase difficulty and degrade performance. As heterogeneity and interaction risk increase, errors rise consistently. In high-density heterogeneous scenes, qualitative inspection indicates that MTR trained on Waymo tends to over-project and under-capture subtle multi-agent interactions, highlighting the challenge of forecasting complex, VRU-rich traffic.

% ---- Palette (once is enough) ----
\definecolor{TrainBar}{RGB}{222,235,255}   % light blue
\definecolor{TrainText}{RGB}{23,79,134}    % dark blue text
\definecolor{TestCol}{RGB}{255,239,213}    % light orange/peach
\definecolor{TestText}{RGB}{156,86,0}      % dark orange text

% ================================
% (A) Cross-dataset main table
% HetroD row uses the same-map setting
% ================================
\begin{table}[ht!]
\centering
\caption{\textbf{Cross-dataset Brier-FDE ($\downarrow$)} for \textbf{MTR}~\cite{mtr} and \textbf{Wayformer}~\cite{wayformer}, two state-of-the-art motion prediction models.
%\ychen{what is MTR and Wayformer? Please specify their goal.}
%
Rows correspond to training datasets, and columns to testing datasets.
The \textbf{HetroD} row adopts the \textit{same-map} setting from the ablation study.
%
%\ychen{Please give a one-sentence insight of your experiments.}
Models trained on Waymo or NuScenes generally over-predict due to heterogeneous traffic complexity.
}
\setlength{\tabcolsep}{3.5pt}
\resizebox{0.82\columnwidth}{!}{%
\begin{tabular}{l>{\columncolor{TestCol}}lcccc}
\toprule
& \cellcolor{white} & \multicolumn{4}{c}{\cellcolor{TrainBar}\textcolor{TrainText}{\textbf{Test}}} \\
\cmidrule(lr){3-6}
& \cellcolor{TestCol}\textcolor{TestText}{\textbf{Train}} & \textbf{NuScenes} & \textbf{Waymo*} & \textbf{SinD} & \textbf{HetroD} \\
\midrule
\multirow{4}{*}{\textbf{MTR}}
& \textbf{NuScenes}           & 2.95  & 10.43 & 5.14  & 6.76 \\
& \textbf{Waymo}              & 4.01  & 2.28  & 4.26  & 6.71 \\
& \textbf{SinD}               & 16.07 & 26.34 & 2.06  & 3.30 \\
& \textbf{HetroD}             & 21.39 & 26.49 & 3.71  & 0.44 \\
\addlinespace[6pt]
\midrule
& \cellcolor{white} & \multicolumn{4}{c}{\cellcolor{TrainBar}\textcolor{TrainText}{\textbf{Test}}} \\
\cmidrule(lr){3-6}
& \cellcolor{TestCol}\textcolor{TestText}{\textbf{Train}} & \textbf{NuScenes} & \textbf{Waymo*} & \textbf{SinD} & \textbf{HetroD} \\
\midrule
\multirow{4}{*}{\textbf{Wayformer}}
& \textbf{NuScenes}           & 2.99  & 8.79  & 5.23  & 9.37 \\
& \textbf{Waymo}              & 2.67  & 2.20  & 3.53  & 10.75 \\
& \textbf{SinD}               & 8.23  & 13.40 & 1.96  & 9.23 \\
& \textbf{HetroD}             & 19.57 & 25.28 & 8.06 & 0.75  \\
\bottomrule
\end{tabular}
}

\vspace{2pt}
{\footnotesize Waymo* uses 30\% of its original training data due to resource constraints.}

\label{tab:cross_dataset_main}
\end{table}

% ================================
% (B) HetroD Ablation Table
% ================================
\begin{table}[!ht]
\centering
\caption{\textbf{Ablation on HetroD under different training/testing splits.}
We compare three regimes: \textit{same-map}, \textit{different-map}, and \textit{different-time}.
Values report Brier-FDE ($\downarrow$) on the HetroD test set.
Parentheses indicate percentage change relative to the in-domain baseline (\textit{same-map}).}
\vspace{3pt}
\setlength{\tabcolsep}{6pt}
\resizebox{0.68\columnwidth}{!}{%
\begin{tabular}{lcc}
\toprule
\textbf{Setting} & \textbf{MTR} & \textbf{Wayformer} \\
\midrule
\textbf{Same-map}  & \textbf{0.44} & \textbf{0.75} \\
\textbf{Diff-map}  & 1.17 (+166\%) & 1.53 (+104\%) \\
\textbf{Diff-time} & 0.42 ($-$5\%) & 0.76 (+1\%) \\
\bottomrule
\end{tabular}
}
\label{tab:hetrod_ablation}
\end{table}

\sisetup{
  table-number-alignment = center,
  table-format = 1.2,        
  detect-weight = true,
  detect-inline-weight = math
}
\definecolor{BestCell}{HTML}{E6F4EA}   
\definecolor{Band}{gray}{0.96}

\begin{table}[!t]
\centering
\caption{\textbf{Scenario-conditioned Brier-FDE ($\downarrow$) on HetroD test scenarios.} MTR models trained on Waymo*, SinD, and HetroD are evaluated on stratified HetroD test cases grouped by agent type, cross-type TTC risk, and scene-level heterogeneity.}
\label{tab:scenario_conditioned}
\vspace{3pt}
\setlength{\tabcolsep}{4pt}

\resizebox{\columnwidth}{!}{%
\begin{tabular}{@{}l *{3}{S} @{}}
\toprule
\textbf{Scenario} & {\textbf{MTR-Waymo*}} & {\textbf{MTR-SinD}} & {\textbf{MTR-HetroD}} \\
\midrule

% -------- Agent type --------
\rowcolor{Band}\multicolumn{4}{@{}l}{\textit{Agent type}}\\[-3pt]
\cmidrule(lr){1-4}
Vehicle                                & 3.64 & 2.55 & 0.83 \\
Two-wheeler                                & 8.69 & 4.63 & 1.16 \\
Pedestrian                             & 2.85 & 1.17 & 0.26 \\
\addlinespace[4pt]

% -------- TTC risk --------
\rowcolor{Band}\multicolumn{4}{@{}l}{\textit{Cross-type TTC risk}}\\[-3pt]
\cmidrule(lr){1-4}
High risk (TTC$<2$\,s)                 & 8.51 & 4.76 & 1.25 \\
Moderate ($2\le \min\mathrm{TTC}<4$\,s)& 8.48 & 4.67 & 1.19 \\
Low (TTC$>4$\,s)                       & 7.87 & 4.00 & 0.90 \\
\addlinespace[4pt]

% -------- Heterogeneity --------
\rowcolor{Band}\multicolumn{4}{@{}l}{\textit{Scene-level heterogeneity}}\\[-3pt]
\cmidrule(lr){1-4}
Low heterogeneity density              & 5.01 & 3.08 & 0.90 \\
High heterogeneity density             & 8.04 & 4.25 & 1.06 \\
\bottomrule
\end{tabular}%
}
\caption*{\footnotesize Per-block \% of filtered HetroD evaluation cases. Agent: Vehicles 39.4\%, Two-wheelers 51.5\%, Pedestrians 9.1\%; TTC: High 26.5\%, Moderate 8.7\%, Low 64.8\%; Scene heterogeneity: Low 66.3\%, High 33.7\%.}

\vspace{-10pt}
    
\end{table}

\subsubsection{Latent Scenario Embeddings}

\begin{figure*}[t]
    \centering
    \includegraphics[width=0.9\textwidth]{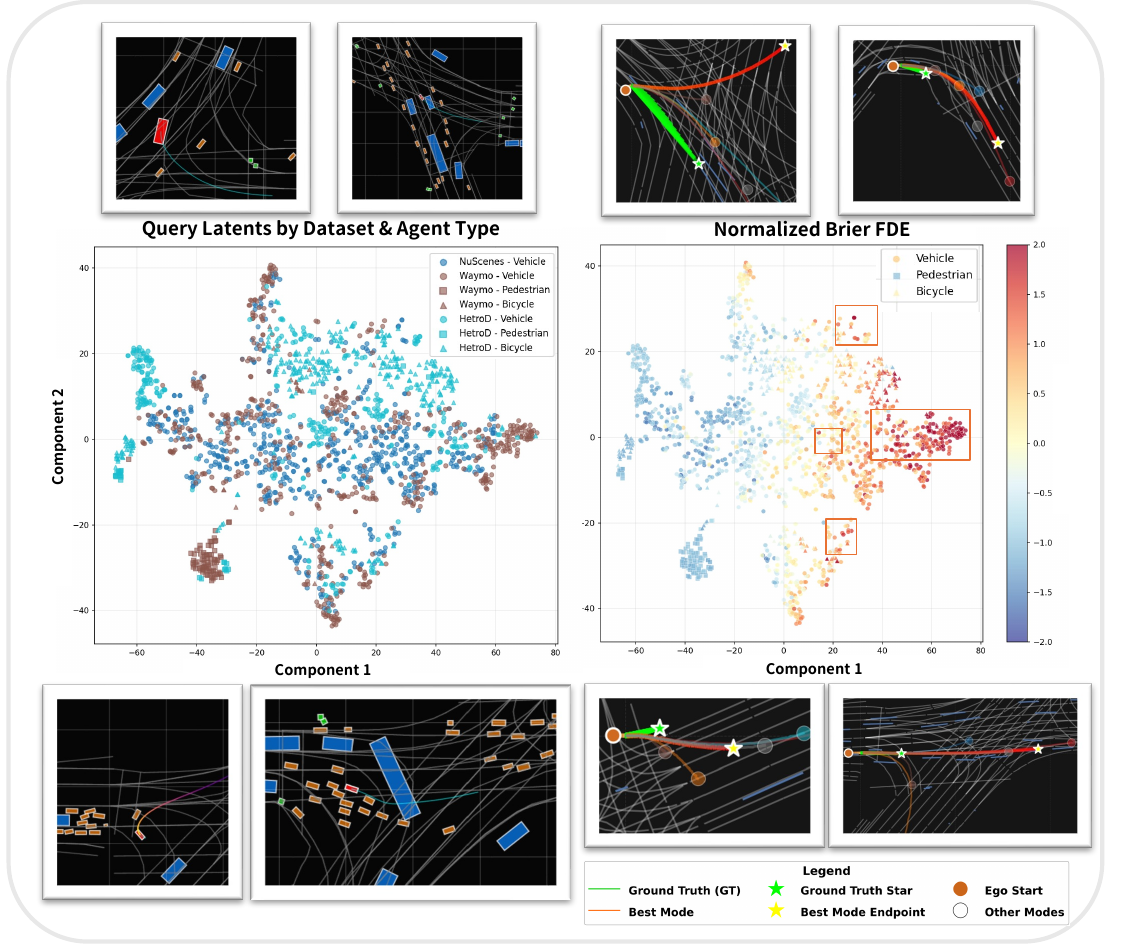}
    \caption{\textbf{Latent scenario embeddings (left) and error map (right).}
        \emph{Left:} Points represent scenarios colored by dataset × agent type.
        \emph{Right:} Same embedding colored by normalized Brier-FDE ($\downarrow$); warmer indicates higher error. Orange boxes highlight characteristic clusters.
        \emph{Left insets:} Example HetroD scenarios with complex behaviors.
        \emph{Right insets:} Model predictions in dense VRU scenarios.
        }
        \vspace{-10pt}
    \label{fig:latent_embedding}
\end{figure*}

We extract the final decoder query embeddings from the Wayformer model trained on Waymo (chosen for its strong cross-domain performance), pool them per scenario, and project to 2D with t-SNE~\cite{tsne} for cross-dataset visualization. As shown in Fig.~\ref{fig:latent_embedding}, randomly sampled scenarios indicate that Waymo and NuScenes share similar latent structure with substantial overlap, whereas HetroD scenarios occupy distinct regions, reflecting complex behaviors and marked differences in heterogeneity, density, and interaction patterns relative to popular on-board datasets.

We further annotate scenarios with their normalized Brier-FDE scores for qualitative analysis. Results show that HetroD scenarios are particularly challenging: the model frequently over-predicts in dense VRU interactions, failing to fully capture rich frontal interactions and nuanced intent among crowded agents. The small physical size of two-wheelers and pedestrians makes anticipating their lateral interactions with the ego agent more difficult. Moreover, unique and complex behaviors (e.g., irregular maneuvers or unexpected lane usage) remain extremely difficult to forecast and generalize.

\subsection{Motion Planning}

We evaluate planner performance using the V-Max framework~\cite{charraut2025vmaxreinforcementlearningframework} under the NuPlan closed-loop, non-reactive scoring protocol. We use 4{,}420 HetroD vehicle scenarios and compare two rule-based planners: the Intelligent Driver Model (IDM)~\cite{idm} and PDM-Closed~\cite{pdm}, a strong planner from the NuPlan benchmark~\cite{nuplan, waymax}. To better reflect the challenges of VRU interactions, we augment the evaluation with a VRU-specific collision breakdown that separates \textit{front} and \textit{lateral} contacts involving two-wheelers and pedestrians, cases that standard forward-collision checks often miss in overtakes and unstructured flows.

\begin{table*}[!ht]
\centering
\caption{\textbf{Closed-loop, non-reactive planning results.} To ensure a fair comparison, we disabled the off-road penalty in the NuPlan aggregate score because, in dense, high-flow, and very narrow-road scenes, these rule-based planners tend to rigidly follow map centerlines rather than adapting, making off-road violations disproportionately likely and obscuring interaction-induced failures.
% \ychen{fonts are too small.} \textcolor{blue}{change to double column}
}
\vspace{3pt}
% \resizebox{.49\textwidth}{!}{%
\begin{tabular}{llcccccc}
\toprule
\textbf{Dataset} & \textbf{Planner} & 
\makecell{\textbf{NuPlan}\\\textbf{Score} ↑} & 
\makecell{\textbf{TTC}\\\textbf{Within Bound} ↑} & 
\makecell{\textbf{Progress}\\\textbf{Ratio} ↑} & 
\makecell{\textbf{Multiple}\\\textbf{Lane Score} ↑} & 
\makecell{\textbf{Comfort} ↑} & 
\makecell{\textbf{At-Fault}\\\textbf{Collisions} ↓} \\
\midrule
\multirow{2}{*}{\textit{NuPlan}}  
  & IDM        & 0.85 & 0.94 & 0.92 & 0.99 & 0.48 & 0.016 \\
  & PDM-Closed & 0.83 & 0.97 & 0.91 & 0.99 & 0.31 & 0.006 \\
\midrule[1pt]
\multirow{2}{*}{\textit{HetroD}} 
  & IDM        & 0.68 & 0.91 & 0.81 & 0.89 & 0.37 & 0.074 \\
  & PDM-Closed & 0.70 & 0.95 & 0.78 & 0.97 & 0.21 & 0.040 \\
\bottomrule
\end{tabular}%
% }
\vspace{-10pt}
    
\label{tab:planner_evaluation}
\end{table*}

\begin{table}[!ht]
\centering
\caption{
\textbf{VRU collision breakdown on \textit{HetroD}.}
At-fault collision rate (↓) decomposed into VRU \emph{front} vs.\ \emph{lateral} contacts for IDM and PDM-Closed.
}
\resizebox{.40\textwidth}{!}{%
\begin{tabular}{lccc}
\toprule
\textbf{Planner} 
& \makecell{\textbf{At-Fault} \\ \textbf{Collision Rate}} 
& \makecell{\textbf{VRU Front} \\ \textbf{Collision Rate}} 
& \makecell{\textbf{VRU Lateral} \\ \textbf{Collision Rate}} \\
\midrule
IDM & 0.074 & 0.008 & 0.031 \\
PDM-Closed & 0.040 & 0.004 & 0.022 \\
\bottomrule
\end{tabular}
}
\vspace{-10pt}
    
\label{tab:hetrd_collision_breakdown}

\end{table}

\begin{figure}[h]
    \centering
    \includegraphics[width=0.48\textwidth]{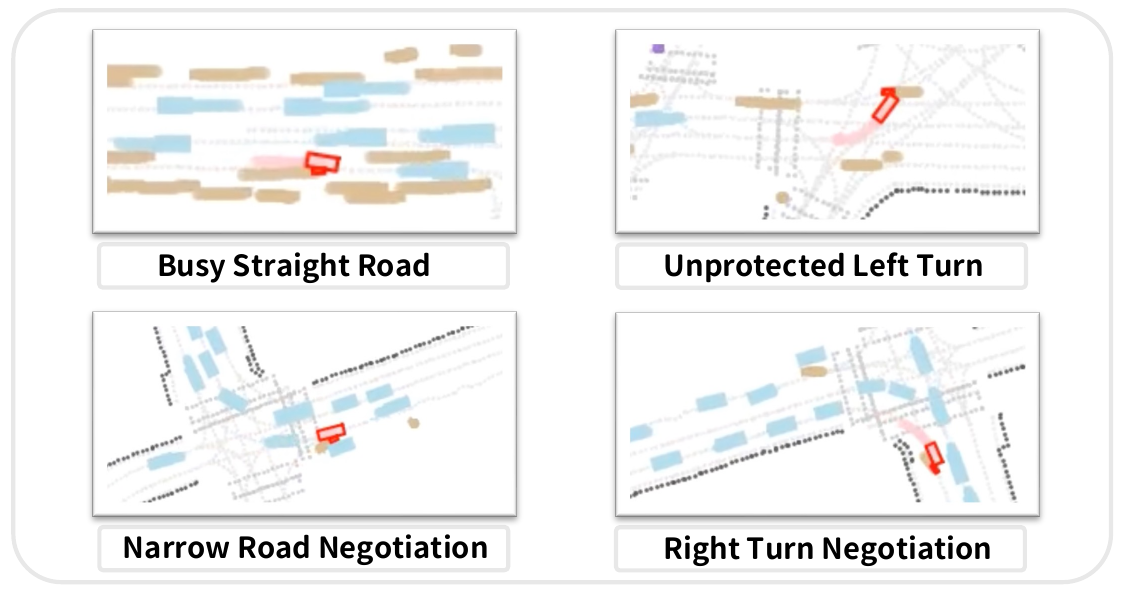}
    \caption{PDM-Closed planner failures on HetroD. Red boxes indicate ego vehicle collisions. Colors denote traffic participants (blue: vehicles, brown: VRUs).}
    \label{fig:planning_qualitative}
\end{figure}

As shown in Table~\ref{tab:planner_evaluation}, both rule-based planners exhibit clear performance drops on HetroD compared to NuPlan: aggregate scores decrease, comfort deteriorates, and at-fault collisions rise. The collision breakdown in Table~\ref{tab:hetrd_collision_breakdown} shows that failures on HetroD frequently involve \textit{lateral} VRU interactions, indicating that planners optimized for structured, vehicle-centric settings struggle to anticipate side interactions. Fig.~\ref{fig:planning_qualitative} illustrates representative failure modes, where PDM-Closed struggles with four common heterogeneous traffic scenarios: (1) busy straight road navigation requiring continuous multi-agent reasoning, (2) unprotected left turns with crossing VRUs, (3) narrow road negotiations demanding precise lateral spacing, and (4) right turn negotiations with surrounding VRUs. These qualitative examples reinforce our quantitative findings—rule-based planners lack the interaction awareness needed for dense, heterogeneous traffic, motivating planning objectives that explicitly account for lateral, multi-agent behaviors in unstructured environments.

\section{Conclusions}
In this work, we introduce HetroD, the first dataset and benchmark that addresses the need for heterogeneous traffic. It comprises 65.4k VRU-rich trajectories, HD maps, and traffic signal states. Our evaluation reveals fundamental limitations in current autonomous driving approaches when deployed in heterogeneous traffic scenarios. As traffic heterogeneity increases, state-of-the-art forecasting and planning methods exhibit significant performance degradation. We identify two critical failure modes: two-wheelers emerge as the most challenging agent type for accurate prediction, while rule-based planners demonstrate disproportionately high lateral VRU collision rates despite maintaining strong lane-keeping performance. Therefore, future work will focus on utilizing HetroD for VRU modeling and simulation, ultimately increasing the safety and compatibility of autonomous driving in heterogeneous traffic environments.

% develop interaction-aware VRUs behavior simulation and leverage occlusion-free aerial coverage for cooperative, heterogeneity-aware planning to improve robustness and transfer across heterogeneous traffic.

% We also notice a strong dataset dependence in existing prediction and planning , highlighting the need for future datasets to emphasize greater diversity and domain coverage

% Our results also suggest that existing methods show a strong dependence on the dataset, indicating that future datasets may benefit from greater diversity and broader domain coverage. 

\section*{Acknowledgment}
This work was supported in part by the National Science and Technology Council under Grants 113-2628-E-A49-022 and 114-2628-E-A49-007, the H\&J Global Chair, the Ministry of Education, and the Yushan Fellow Program Administrative Support Grant. W.J. Chang was also supported by the National Science Foundation Graduate Research Fellowship Program under Grant No. DGE-2146752. Any opinions, findings, and conclusions or recommendations expressed in this material are those of the author(s) and do not necessarily reflect the views of the National Science Foundation.

\bibliographystyle{IEEEtran}
\bibliography{root.bib}

@INPROCEEDINGS{SIND,  author={Xu, Yanchao and Shao, Wenbo and Li, Jun and Yang, Kai and Wang, Weida and Huang, Hua and Lv, Chen and Wang, Hong},  
booktitle={ITSC},   
title={{SIND: A Drone Dataset at Signalized Intersection in China}},  
year={2022}}

@inproceedings{dsc3d,
  title        = {{Highly Accurate and Diverse Traffic Data: The DeepScenario Open 3D Dataset}},
  author       = {Dhaouadi,Oussema and Meier, Johannes and Wahl, Luca and Kaiser, Jacques and Scalerandi, Luca and Wandelburg, Nick and Zhuo, Zhuolun and Berinpanathan, Nijanthan and Banzhaf, Holger and Cremers, Daniel},
  booktitle    = {IV},
  year         = {2025},
}

@misc{charraut2025vmaxreinforcementlearningframework,
      title={{V-Max: A Reinforcement Learning Framework for Autonomous Driving}},
      author={Valentin Charraut and Thomas Tournaire and Waël Doulazmi and Thibault Buhet},
      year={2025},
      booktitle    = {arXiv preprint},
    
}

@article{INTERACTION,
title = {{INTERACTION} {Dataset}: {An} {INTERnational}, {Adversarial} and {Cooperative} {moTION} {Dataset} in {Interactive} {Driving} {Scenarios} with {Semantic} {Maps}},
journal = {arXiv:1910.03088 [cs, eess]},
author = {Zhan, Wei and Sun, Liting and Wang, Di and Shi, Haojie and Clausse, Aubrey and Naumann, Maximilian and K\"ummerle, Julius and K\"onigshof, Hendrik and Stiller, Christoph and de La Fortelle, Arnaud and Tomizuka, Masayoshi},
year = {2019}}

@inproceedings{highD,
  author       = {Robert Krajewski and
                  Julian Bock and
                  Laurent Kloeker and
                  Lutz Eckstein},
  title        = {{The highD Dataset: {A} Drone Dataset of Naturalistic Vehicle Trajectories
                  on German Highways for Validation of Highly Automated Driving Systems}},
  booktitle    = {ITSC},
  year         = {2018}
}

@inproceedings{Automatum,
  author       = {Paul Spannaus and
                  Peter Zechel and
                  Kilian Lenz},
  title        = {{{AUTOMATUM} {DATA:} Drone-based highway dataset for the development
                  and validation of automated driving software for research and commercial
                  applications}},
  booktitle    = {IV},
  year         = {2021}
}

@inproceedings{exID,
  author       = {Tobias Moers and
                  Lennart Vater and
                  Robert Krajewski and
                  Julian Bock and
                  Adrian Zlocki and
                  Lutz Eckstein},
  title        = {{The exiD Dataset: {A} Real-World Trajectory Dataset of Highly Interactive
                  Highway Scenarios in Germany}},
  booktitle    = {IV},
  year         = {2022}
}

@inproceedings{openDD,
  author       = {Antonia Breuer and
                  Jan{-}Aike Term{\"{o}}hlen and
                  Silviu Homoceanu and
                  Tim Fingscheidt},
  title        = {{openDD: A Large-Scale Roundabout Drone Dataset}},
  booktitle    = {ITSC},
  year         = {2020}
}

@inproceedings{CITR_DUT,
  author       = {Dongfang Yang and
                  Linhui Li and
                  Keith A. Redmill and
                  {\"{U}}mit {\"{O}}zg{\"{u}}ner},
  title        = {{Top-view Trajectories: A Pedestrian Dataset of Vehicle-Crowd Interaction
                  from Controlled Experiments and Crowded Campus}},
  booktitle    = {IV},
  year         = {2019}
}

@inproceedings{Stanford_Drone,
  author       = {Alexandre Robicquet and
                  Amir Sadeghian and
                  Alexandre Alahi and
                  Silvio Savarese},
  title        = {{Learning Social Etiquette: Human Trajectory Understanding In Crowded
                  Scenes}},
  booktitle    = {ECCV},
  year         = {2016}
}

@inproceedings{CTV,
  author       = {Awad Mukbil and
                  Yasin Maan Yousif and
                  Sakif Hossain and
                  J{\"{o}}rg P. M{\"{u}}ller},
  title        = {{CTV-Dataset: A Shared Space Drone Dataset for Cyclist-Road User
                  Interaction Derived from Campus Experiments}},
  booktitle    = {ITSC},
  year         = {2023}
}

@inproceedings{inD,
    title={{The inD Dataset: A Drone Dataset of Naturalistic Road User Trajectories at German Intersections}},
    author={Bock, Julian and Krajewski, Robert and Moers, Tobias and Runde, Steffen and Vater, Lennart and Eckstein, Lutz},
    booktitle={IV},
    year={2020},
}

@article{JKU_DORA,
  author       = {Pavlo Tkachenko and
                  Novel Certad and
                  Gunda Singer and
                  Cristina Olaverri{-}Monreal and
                  Luigi del Re},
  title        = {{The JKU DORA Traffic Dataset}},
  journal      = {{IEEE} Access},
  year         = {2022}
}

@inproceedings{GroundMix,
  author    = {Meier, Johannes and Scalerandi, Luca and Dhaouadi, Oussema and Kaiser, Jacques and Araslanov Nikita and Cremers, Daniel},
  title     = {{CARLA Drone: Monocular 3D Object Detection from a Different Perspective}},
  booktitle   = {GCPR},
  year      = {2024}
}

@inproceedings{rounD,
  author       = {Robert Krajewski and
                  Tobias Moers and
                  Julian Bock and
                  Lennart Vater and
                  Lutz Eckstein},
  title        = {{The rounD Dataset: A Drone Dataset of Road User Trajectories at
                  Roundabouts in Germany}},
  booktitle    = {ITSC},
  year         = {2020}
}

@article{CitySim,
author = {Ou Zheng and Mohamed Abdel-Aty and Lishengsa Yue and Amr Abdelraouf and Zijin Wang and Nada Mahmoud},
title ={{CitySim: A Drone-Based Vehicle Trajectory Dataset for Safety-Oriented Research and Digital Twins}},
journal = {Transportation Research Record},
year = {2024}
}

@article{AD4CHE,
  author       = {Yuxin Zhang and
                  Cheng Wang and
                  Ruilin Yu and
                  Luyao Wang and
                  Wei Quan and
                  Yang Gao and
                  Pengfei Li},
  title        = {{The AD4CHE Dataset and Its Application in Typical Congestion Scenarios
                  of Traffic Jam Pilot Systems}},
  year         = {2023},
 journal={IEEE Trans. Intell. Veh.}, 
}

@inproceedings{WOMD,
  author       = {Kan Chen and
                  Runzhou Ge and
                  Hang Qiu and
                  Rami Ai{-}Rfou and
                  Charles R. Qi and
                  Xuanyu Zhou and
                  Zoey Yang and
                  Scott Ettinger and
                  Pei Sun and
                  Zhaoqi Leng and
                  Mustafa Baniodeh and
                  Ivan Bogun and
                  Weiyue Wang and
                  Mingxing Tan and
                  Dragomir Anguelov},
  title        = {{WOMD-LiDAR: Raw Sensor Dataset Benchmark for Motion Forecasting}},
  booktitle    = {CORR},
  year         = {2023}
}

@INPROCEEDINGS{lanelet2,
  author={Poggenhans, Fabian and Pauls, Jan-Hendrik and Janosovits, Johannes and Orf, Stefan and Naumann, Maximilian and Kuhnt, Florian and Mayr, Matthias},
  booktitle={ITSC}, 
  title={{Lanelet2: A high-definition map framework for the future of automated driving}}, 
  year={2018},
}

@inproceedings{Waymo, 
author       = {Pei Sun and
                  Henrik Kretzschmar and
                  Xerxes Dotiwalla and
                  Aurelien Chouard and
                  Vijaysai Patnaik and
                  Paul Tsui and
                  James Guo and
                  Yin Zhou and
                  Yuning Chai and
                  Benjamin Caine and
                  Vijay Vasudevan and
                  Wei Han and
                  Jiquan Ngiam and
                  Hang Zhao and
                  Aleksei Timofeev and
                  Scott Ettinger and
                  Maxim Krivokon and
                  Amy Gao and
                  Aditya Joshi and
                  Yu Zhang and
                  Jonathon Shlens and
                  Zhifeng Chen and
                  Dragomir Anguelov},
  title        = {{Scalability in Perception for Autonomous Driving: Waymo Open Dataset}},
  booktitle    = {CVPR},
  year         = {2020},
}

@INPROCEEDINGS{nuScenes,
  title={{nuScenes: A multimodal dataset for autonomous driving}},
  author={Holger Caesar and Varun Bankiti and Alex H. Lang and Sourabh Vora and 
          Venice Erin Liong and Qiang Xu and Anush Krishnan and Yu Pan and 
          Giancarlo Baldan and Oscar Beijbom}, 
  booktitle={CVPR},
  year=2020
}

@misc{V2AIX,
      title={{V2AIX: A Multi-Modal Real-World Dataset of ETSI ITS V2X Messages in Public Road Traffic}}, 
      author={Guido Kueppers and Jean-Pierre Busch and Lennart Reiher and Lutz Eckstein},
      year={2024},
      booktitle    = {arXiv preprint},
}

@inproceedings{V2X-Seq,
  title={{V2X-Seq: A large-scale sequential dataset for vehicle-infrastructure cooperative perception and forecasting}},
  author={Yu, Haibao and Yang, Wenxian and Ruan, Hongzhi and Yang, Zhenwei and Tang, Yingjuan and Gao, Xu and Hao, Xin and Shi, Yifeng and Pan, Yifeng and Sun, Ning and Song, Juan and Yuan, Jirui and Luo, Ping and Nie, Zaiqing},
  booktitle={CVPR},
  year={2023},
}

@inproceedings{MONA,
  author       = {Luis Gressenbuch and
                  Klemens Esterle and
                  Tobias Kessler and
                  Matthias Althoff},
  title        = {{MONA: The Munich Motion Dataset of Natural Driving}},
  booktitle    = {ITSC},
  year         = {2022}
}

@INPROCEEDINGS { Argoverse2,
  author = {Benjamin Wilson and William Qi and Tanmay Agarwal and John Lambert and Jagjeet Singh and Siddhesh Khandelwal and Bowen Pan and Ratnesh Kumar and Andrew Hartnett and Jhony Kaesemodel Pontes and Deva Ramanan and Peter Carr and James Hays},
  title = {{Argoverse 2: Next Generation Datasets for Self-driving Perception and Forecasting}},
  booktitle = {NeurIPS},
  year = {2021}
}

@inproceedings{chandra2022meteoradenseheterogeneousunstructured,
      title={{METEOR:A Dense, Heterogeneous, and Unstructured Traffic Dataset With Rare Behaviors}},
      author={Rohan Chandra and Xijun Wang and Mridul Mahajan and Rahul Kala and Rishitha Palugulla and Chandrababu Naidu and Alok Jain and Dinesh Manocha},
      year={2023},
      booktitle    = {ICRA},

}

@misc{NGSIM,
	author = {U. S. Department of Transportation Federal Highway Administration},
	title = {{N}ext {G}eneration {S}imulation ({N}{G}{S}{I}{M}) {V}ehicle {T}rajectories and {S}upporting {D}ata},
	year = {2016},
        booktitle    = {arXiv preprint},
}

@inproceedings{KITTI,
  author       = {Andreas Geiger and
                  Philip Lenz and
                  Raquel Urtasun},
  title        = {{Are we ready for autonomous driving? The KITTI vision benchmark
                  suite}},
  booktitle    = {CVPR},
  year         = {2012},
}

@article{tsne,
  author  = {Laurens van der Maaten and Geoffrey Hinton},
  title   = {{Visualizing Data using t-SNE}},
  journal = {Journal of Machine Learning Research},
  year    = {2008},
}

@book{shannon,
	title = {{The mathematical theory of communication}},
	publisher = {University of Illinois Press},
	author = {Shannon, Claude E. and Weaver, Warren},
	year = {1949},
}

@misc{opendrive,
  title={{OpenDRIVE: Open Dynamic Road Information for Vehicle Environment}},
  author={{OpenDRIVE Initiative}},
  year={2006},
}

@article{kalmen_filter,
  title={{A new approach to linear filtering and prediction problems}},
  author={Kalman, Rudolph Emil},
  journal={Journal of basic Engineering},
  year={1960},
}

@inproceedings{lyft,
  author       = {John Houston and
                  Guido Zuidhof and
                  Luca Bergamini and
                  Yawei Ye and
                  Long Chen and
                  Ashesh Jain and
                  Sammy Omari and
                  Vladimir Iglovikov and
                  Peter Ondruska},
  title        = {{One Thousand and One Hours: Self-driving Motion Prediction Dataset}},
  booktitle    = {CoRL},
  year         = {2020}
}

@article{v2xreal,
  title={{V2X-Real: a Large-Scale Dataset for Vehicle-to-Everything Cooperative Perception}},
  author={Xiang, Hao and Zheng, Zhaoliang and Xia, Xin and Xu, Runsheng and Gao, Letian and Zhou, Zewei and Han, Xu and Ji, Xinkai and Li, Mingxi and Meng, Zonglin and others},
  journal={arXiv preprint arXiv:2403.16034},
  year={2024}
}

@article{tumtraf,
    title={{TUMTraf V2X Cooperative Perception Dataset}},
    author={Zimmer, Walter and Wardana, Gerhard Arya and Sritharan, Suren and Zhou, Xingcheng and Song, Rui and Knoll, Alois},
    journal={arXiv preprint arXiv:2403.01316},
    year={2024}
}

@article{scenarionet,
  title={{ScenarioNet: Open-Source Platform for Large-Scale Traffic Scenario Simulation and Modeling}},
  author={Li, Quanyi and Peng, Zhenghao and Feng, Lan and Liu, Zhizheng and Duan, Chenda and Mo, Wenjie and Zhou, Bolei},
  journal={NeurIPS},
  year={2023}
}

@article{metadrive,
  title={{Metadrive: Composing diverse driving scenarios for generalizable reinforcement learning}},
  author={Li, Quanyi and Peng, Zhenghao and Feng, Lan and Zhang, Qihang and Xue, Zhenghai and Zhou, Bolei},
  journal={TPAMI},
  year={2022}
}

@article{unitraj,
  title={{UniTraj: A Unified Framework for Scalable Vehicle Trajectory Prediction}},
  author={Feng, Lan and Bahari, Mohammadhossein and Amor, Kaouther Messaoud Ben and Zablocki, {\'E}loi and Cord, Matthieu and Alahi, Alexandre},
  journal={arXiv preprint arXiv:2403.15098},
  year={2024}
}

@inproceedings{gpudrive,
      title={{GPUDrive: Data-driven, multi-agent driving simulation at 1 million FPS}},
      author={Saman Kazemkhani and Aarav Pandya and Daphne Cornelisse and Brennan Shacklett and Eugene Vinitsky},
      booktitle={ICLR},
      year={2025},
}

@inproceedings{dronalize,
  title={{Toward Unified Practices in Trajectory Prediction Research on Bird's-Eye-View Datasets}},
  author={Westny, Theodor and Olofsson, Bj{\"o}rn and Frisk, Erik},
  booktitle={IV},
  year={2025}
}

@Inproceedings{trajdata,
  author = {Ivanovic, Boris and Song, Guanyu and Gilitschenski, Igor and Pavone, Marco},
  title = {{trajdata: A Unified Interface to Multiple Human Trajectory Datasets}},
  booktitle = {{NeurIPS}},
  year = {2023},
}

@INPROCEEDINGS{nuplan, 
  title={{NuPlan: A closed-loop ML-based planning benchmark for autonomous vehicles}},
  author={H. Caesar, J. Kabzan, K. Tan et al.},
  booktitle={CVPR ADP3 workshop},
  year=2021
}

@inproceedings{waymax, title={{Waymax: An Accelerated, Data-Driven Simulator for Large-Scale Autonomous Driving Research}}, author={Gulino, Cole and Fu, Justin and Luo, Wenjie and Tucker, George and Bronstein, Eli and Lu, Yiren and Harb, Jean and Pan, Xinlei and Wang, Yan and Chen, Xiangyu and Co-Reyes, John D and Agarwal, Rishabh and Roelofs, Rebecca and Lu, Yao and Montali, Nico and Mougin, Paul and Yang, Zoey and White, Brandyn and Faust, Aleksandra and McAllister, Rowan and Anguelov, Dragomir and Sapp, Benjami}, booktitle={NeurIPS},year={2023}}

@article{mtr,
  title={{Motion transformer with global intention localization and local movement refinement}},
  author={Shi, Shaoshuai and Jiang, Li and Dai, Dengxin and Schiele, Bernt},
  journal={NeurIPS},
  year={2022}
}

@article{wayformer,
  title={{Wayformer: Motion Forecasting via Simple \& Efficient Attention Networks}},
  author={Nigamaa Nayakanti and Rami Al-Rfou and Aurick Zhou and Kratarth Goel and Khaled S. Refaat and Benjamin Sapp},
  journal={ICRA},
  year={2022},
}

@InProceedings{pdm,
  title={{Parting with Misconceptions about Learning-based Vehicle Motion Planning}},
  author={Dauner, Daniel and Hallgarten, Marcel and Geiger, Andreas and Chitta, Kashyap},
  booktitle={CoRL},
  year={2023}
}

@inproceedings{TraPHic,
   title={{TraPHic: Trajectory Prediction in Dense and Heterogeneous Traffic Using Weighted Interactions}},
   booktitle={CVPR},
   author={Chandra, Rohan and Bhattacharya, Uttaran and Bera, Aniket and Manocha, Dinesh},
   year={2019}, }

@ARTICLE{surveyondata,
  author={Liu, Mingyu and Yurtsever, Ekim and Fossaert, Jonathan and Zhou, Xingcheng and Zimmer, Walter and Cui, Yuning and Zagar, Bare Luka and Knoll, Alois C.},
  journal={IEEE Trans. Intell. Veh.}, 
  title={{A Survey on Autonomous Driving Datasets: Statistics, Annotation Quality, and a Future Outlook}}, 
  year={2024},}

@inproceedings{pandaset,
      title={{PandaSet: Advanced Sensor Suite Dataset for Autonomous Driving}},
      author={Xiao, Pengchuan and Shao, Zhenlei and Hao, Steven and Zhang, Zishuo and Chai, Xiaolin and Jiao, Judy and Li, Zesong and Wu, Jian and Sun, Kai and Jiang, Kun and Wang, Yunlong and Yang, Diange},
      year={2021},
      booktitle = {ITSC},
}

@article{bdd100k,
  title={{BDD100K: A Diverse Driving Dataset for Heterogeneous Multitask Learning}},
  author={Fisher Yu and Haofeng Chen and Xin Wang and Wenqi Xian and Yingying Chen and Fangchen Liu and Vashisht Madhavan and Trevor Darrell},
  journal={CVPR},
  year={2018},
}

@inproceedings{h3d,
    author = {Abhishek Patil and Srikanth Malla and Haiming Gang and Yi-Ting Chen},
    title = {{The H3D Dataset for Full-Surround 3D Multi-Object Detection and Tracking in Crowded Urban Scenes}},  
    booktitle = {ICRA},
    year = {2019}
}

@inproceedings{dairv2x,
  title={{Dair-v2x: A large-scale dataset for vehicle-infrastructure cooperative 3d object detection}},
  author={Yu, Haibao and Luo, Yizhen and Shu, Mao and Huo, Yiyi and Yang, Zebang and Shi, Yifeng and Guo, Zhenglong and Li, Hanyu and Hu, Xing and Yuan, Jirui and Nie, Zaiqing},
  booktitle={CVPR},
  year={2022}
}

@inproceedings{opv2v,
      title={{OPV2V: An Open Benchmark Dataset and Fusion Pipeline for Perception with Vehicle-to-Vehicle Communication}},
      author={Runsheng Xu and Hao Xiang and Xin Xia and Xu Han and Jinlong Li and Jiaqi Ma},
      year={2022},
      booktitle = {ICRA},
}

@ARTICLE{ttc,
	author = {Ward, James R. and Agamennoni, Gabriel and Worrall, Stewart and Bender, Asher and Nebot, Eduardo},
	title = {{Extending Time to Collision for probabilistic reasoning in general traffic scenarios}},
	year = {2015},
	journal = {Transportation Research Part C: Emerging Technologies},
}

@Article{drac,
AUTHOR = {Shen, Jiajun and Yang, Guangchuan},
TITLE = {{Crash Risk Assessment for Heterogeneity Traffic and Different Vehicle-Following Patterns Using Microscopic Traffic Flow Data}},
JOURNAL = {Sustainability},
YEAR = {2020},
}

@article{idm,
   title={{Congested traffic states in empirical observations and microscopic simulations}},
   journal={Physical Review E},
   author={Treiber, Martin and Hennecke, Ansgar and Helbing, Dirk},
   year={2000}, }

@article{SDDComplex,
author = {Andle, Joshua and Soucy, Nicholas and Socolow, Simon and Yasaei Sekeh, Salimeh},
year = {2022},
title = {{The Stanford Drone Dataset Is More Complex Than We Think: An Analysis of Key Characteristics}},
journal = {IEEE Trans. Intell. Veh.},
}

@Article{carla+,
AUTHOR = {Malik, Sumbal and Khan, Manzoor Ahmed and Aadam and El-Sayed, Hesham and Iqbal, Farkhund and Khan, Jalal and Ullah, Obaid},
TITLE = {{CARLA+: An Evolution of the CARLA Simulator for Complex Environment Using a Probabilistic Graphical Model}},
JOURNAL = {Drones},
YEAR = {2023},
}

@article{Accid3nD,
  title={{Towards Vision Zero: The Accid3nD Dataset}},
  author={Zimmer, Walter and Greer, Ross and Lehmberg, Daniel and Pavel, Marc and Caesar, Holger and Zhou, Xingcheng and Ghita, Ahmed and Trivedi, Mohan and Song, Rui and Cao, Hu and others},
  journal={arXiv preprint arXiv:2503.12095},
  year={2025}
}

\end{document}